\documentclass[acmconf,screen,nonacm,review=false,timestamp=false]{acmart}

\usepackage{caption}
\usepackage{subcaption}
\usepackage[capitalize,noabbrev]{cleveref}  

\newcommand{\EE}{\mathbb{E}}

\newcommand\blfootnote[1]{%
  \begingroup
  \renewcommand\thefootnote{}\footnote{#1}%
  \addtocounter{footnote}{-1}%
  \endgroup
}

\AtBeginDocument{%
  }

\setcopyright{acmcopyright}
\copyrightyear{2023}
\acmYear{2023}

\acmConference[CONSEQUENCES '23]{Make sure to enter the correct
  conference title from your rights confirmation email}{September 18--22,
  2023}{Singapore}

\begin{document}

\title{Double Clipping: Less-Biased Variance Reduction in Off-Policy Evaluation}

\author{Jan Malte Lichtenberg}
\authornote{Corresponding author, jlichten@amazon.com}
\email{jlichten@amazon.com}


\author{Alexander Buchholz}

\author{Giuseppe Di Benedetto}

\author{Matteo Ruffini}

\author{Ben London}
\email{blondon@amazon.com}
\affiliation{%
  \institution{Amazon Music}
  \country{USA}
}

\renewcommand{\shortauthors}{Lichtenberg et al.}

\begin{abstract}
``Clipping'' (a.k.a. importance weight truncation) is a widely used variance-reduction technique for counterfactual off-policy estimators. Like other variance-reduction techniques, clipping reduces variance at the cost of increased bias. However, unlike other techniques, the bias introduced by clipping is always a downward bias (assuming non-negative rewards), yielding a lower bound on the true expected reward. In this work we propose a simple extension, called \emph{double clipping}, which aims to compensate this downward bias and thus reduce the overall bias, while maintaining the variance reduction properties of the original estimator.
\end{abstract}





\maketitle

\section{Introduction}
\blfootnote{\newline Presented at CONSEQUENCES '23 workshop at RecSys 2023 conference, Singapore.}
Off-policy evaluators are a crucial component in the development of many real-world recommender systems. They allow us to estimate the performance of a new \emph{target} recommendation policy based on interaction data logged from a different \emph{logging} policy (for instance, the current production recommender), thereby reducing the need to run slow and costly A/B tests.

Many counterfactual off-policy estimators are based on the inverse propensity scoring (IPS) principle~\citep{ionides2008truncated, strehl2010learning, bottou2013counterfactual, imbens2015causal}. Given a stochastic logging policy and some mild assumptions, IPS-based estimators are unbiased, but often suffer from high variance. This is true even on industrial-scale data sizes; in particular, if the logging policy is close to being deterministic. Intuitively speaking, most IPS estimators contain propensity ratio weights of the form $w = p_\text{target} / p_\text{logging}$, where $p_\text{target}$ is a target propensity (e.g., the probability that the target policy recommends a particular action to the user) and $p_\text{logging}$ is the logging propensity (e.g., the probability that the logging policy recommended that same action to the user). These ratios can become arbitrarily large for small logging propensities, which then leads to high variance in the overall estimate.

The literature has proposed various variance-reduction techniques for IPS-style estimators, including weight clipping~\citep{ionides2008truncated, bembom2008data, strehl2010learning, bottou2013counterfactual}, self-normalization~\citep{swaminathan2015self}, doubly-robust estimators~\citep{dudik2011doubly, su2020doubly, oosterhuis2023doubly}, as well as generalizations of those ideas~\citep{wang2017optimal, su2019cab, buchholz2022off}. In this article we revisit weight clipping, which is still used extensively due to its simplicity (it does not require a reward model) and its generality (it is readily applicable to IPS-style estimators used in more complex real-world applications, such as ranking~\citep{li2018offline, buchholz2022off} or slate recommendation~\citep{swaminathan2017off}, where self-normalized or doubly-robust estimators are not available or difficult to implement). 

\begin{figure}[t!] 
  \label{fig:exp}
  \centering
  \begin{subfigure}[b]{0.43\textwidth}
    \centering
    \includegraphics[width=\textwidth]{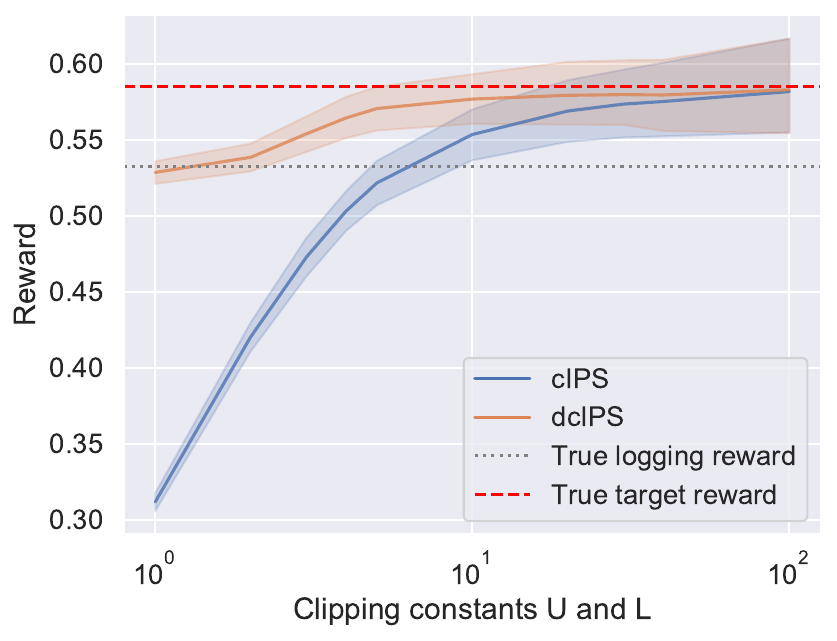}
    \caption{}
    \label{fig:clipping_path}
  \end{subfigure}
  \begin{subfigure}[b]{0.43\textwidth}
    \centering
    \includegraphics[width=\textwidth]{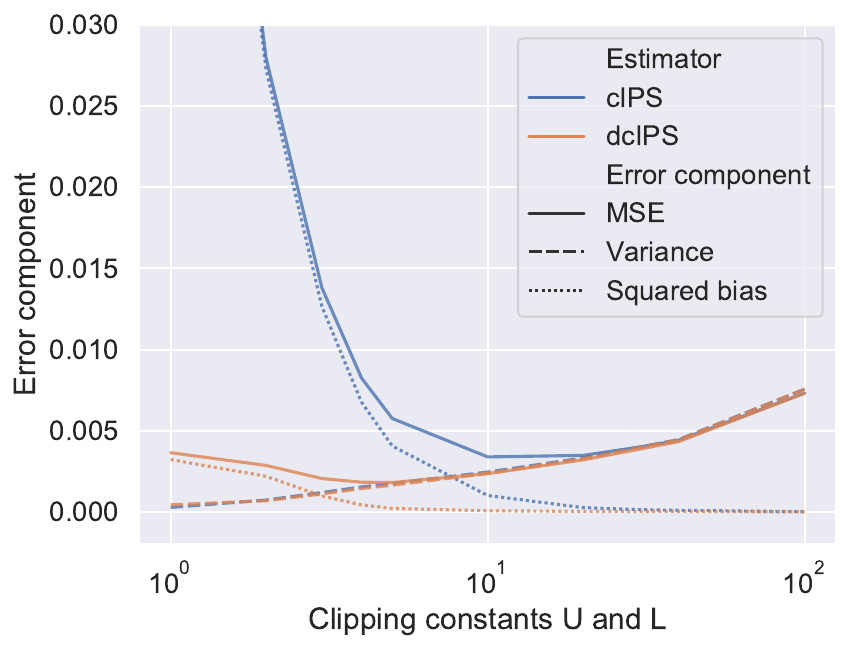}
    \caption{}
    \label{fig:mse}
  \end{subfigure}
  \caption{Comparison of clipped IPS (cIPS, blue) and doubly-clipped IPS (dcIPS, orange) in a synthetic bandit experiment (detailed setup described in Section \ref{sec:exp}). Figure \ref{fig:clipping_path}: Mean (solid line) and corresponding standard error bands of reward estimates across 100 repetitions as a function of clipping constants $U$ (for both cIPS and dcIPS) and $L=U$ (only for dcIPS). The dashed red line shows the true reward of the target policy, i.e., the estimation target. The dotted grey line shows the average logging reward observed in the data set. Figure \ref{fig:mse}: mean squared error (MSE) between reward estimate and true target reward across 100 repetitions. The dashed lines show the variance components, the dotted lines show the squared-bias components for both estimators.}
\end{figure}

The basic idea of weight clipping is to simply avoid large propensity weight ratios by (hard-)clipping the ratios by a constant upper bound $U$, which is usually treated as a hyper-parameter for the estimation procedure. Just like other variance-reduction techniques, the clipping procedure effectively reduces the variance of the IPS estimator at the cost of introducing a bias. Unlike other techniques, however, the bias introduced by clipping is always pessimistic. In other words, on average, the estimator underestimates the true expected reward (under the technical assumption that rewards are always non-negative), as illustrated in Figure \ref{fig:clipping_path}. 

In this work, we exploit this property of the clipping bias, so as to obtain more accurate estimates. Specifically, we clip the propensity ratios from both sides rather than just from above, thereby potentially correcting pessimistic underestimates with optimistic overestimates. Experiments with synthetic data show that this approach leads to a reduction in MSE.


\section{Background and related work}
We focus on off-policy estimation in the standard contextual multi-armed bandit setting, but we note that our work is applicable to counterfactual learning-to-rank~\citep{joachims2017unbiased, li2018offline} and slate recommendation~\citep{swaminathan2017off}.

\subsection{Off-policy evaluation in the contextual bandit.} Consider a contextual bandit setting, where a stochastic \emph{logging policy} $\pi_0(y|x)$ (e.g., the currently deployed recommender system) repeatedly selects an \emph{action} $y \in \mathcal{Y}$ based on a given \emph{context} $x \sim P(\mathcal{X})$ (e.g., user history, action features, etc.). The system then observes a non-negative \emph{reward} $r \sim R(x,y) \geq 0$, which depends on the action that was selected and the context. The system does not observe the rewards for any  action that was not selected by the logging policy. After $n$ rounds, the logged data set is given by $\mathcal{D} = \{(x_i, y_i, r_i, \pi_0(\cdot|x_i))\}$, where $r_i$ is the observed reward and  $\pi_0(y_i|x_i)$ is the \emph{propensity} (i.e., probability) of action $y_i$ to be selected by the logging policy for context $x_i$. The goal of off-policy evaluation is to estimate the expected reward of a new \emph{target} policy $\pi$, given by $R(\pi) = \EE_{x \sim P(\mathcal{X})}\EE_{y \sim \pi(\cdot|x)}\EE_{r \sim R(x, y|x)}[r]$, 
based on the logging data set $\mathcal{D}$. 
The challenge is that the logged data only contains rewards for actions selected by $\pi_0$, which may be different from those selected by $\pi$. Thus, we are faced with a counterfactual estimation problem.

\subsection{Counterfactual off-policy estimators.} The standard inverse propensity scoring (IPS)~\citep{horvitz1952generalization, strehl2010learning, bottou2013counterfactual, swaminathan2015batch} estimate for the contextual bandit problem is given by
\begin{equation} \label{eq:IPS}
    \hat{R}_{\text{IPS}}(\pi) = \frac{1}{n} \sum_{i=1}^n r(x_i, y_i) \frac{\pi(y_i|x_i)}{\pi_0(y_i|x_i)} = \frac{1}{n} \sum_{i=1}^n r(x_i, y_i) w(x_i, y_i).
\end{equation}
The estimator $\hat{R}_{\text{IPS}}(\pi)$ is an unbiased estimator of $R(\pi)$, given the overlap assumption: $\pi_0(y|x) > 0$ whenever $\pi(y|x) > 0$. To satisfy the overlap assumption, the logging policy is usually randomized, leading to the following dilemma. Too much randomization can degrade user experience, but too little randomization leads to high variance in IPS off-policy estimation: little randomization means that some propensity values $\pi_0(y_i|x_i)$ are tiny, which in turn leads to the occasional huge weighting factor $w(x_i, y_i)$. 



A widely used technique to reduce the variance of the standard IPS estimator is to simply clip (some authors also say ``truncate'' or ``trim'') large importance weight ratios. Specifically, we use the clipped IPS estimator (cIPS) that clips the entire ratio, that is, 
\begin{equation} \label{eq:cIPS}
\hat{R}_{\text{cIPS}}(\pi, U)= \frac{1}{n} \sum_{i=1}^n r(x_i, y_i) { \min\{ }w(x_i, y_i), {{U \}}},
\end{equation}
where ${U} \geq 1$ is the \emph{upper clipping constant}.

\section{Clipped IPS is always downward biased.}
 Clearly, for $U=\infty$, the clipped IPS estimator (Eq. \ref{eq:cIPS}) is equivalent to the un-clipped IPS estimator from Eq. \ref{eq:IPS}. With non-negative rewards, decreasing $U$ effectively reduces variance, at the cost of a downward bias,  as illustrated in Figure \ref{fig:clipping_path}. The following proposition confirms this intuition about the downward bias.
\begin{proposition} \label{prop:bias_cIPS}
Let $w(x, y) > 0 ~ \forall x, y$, then 
the bias of $\hat{R}_{\text{cIPS}}(\pi, U)$ is given by (proof in the Appendix)
\begin{equation} \label{eq:biasCIPS}
    \text{Bias}(\hat{R}_{\text{cIPS}}(\pi, U)) = \mathbb{E}_x\mathbb{E}_{y\sim\pi}\Bigl[\underbrace{\vphantom{\frac{U}{w(x, y)}}\mathbf{1}_{\{w(x, y) > U\}}
    }_{\text{Only clipped records}} 
    \underbrace{\left(\frac{U}{w(x, y)}-1\right)}_{\text{Always < 0}}
    \underbrace{\vphantom{\frac{U}{w(x, y)}}\mathbb{E}_r[r(x, y) | x, y]}_{\text{Expected reward}}\Bigr].
\end{equation}
\end{proposition} 
\noindent If the clipping constant $U$ is higher than the highest attainable propensity weight ratio $w(x, y)$ across all requests, then the clipped IPS estimator essentially becomes the standard, unbiased IPS estimator. As soon as the clipping constant becomes ``active'' in the sense that it starts clipping propensity weight ratios, then the bias is always strictly negative assuming non-negative rewards (ignoring the trivial case in which all clipped requests have zero expected reward).

Many machine learning practitioners are happy to accept a small bias to reduce the variance of their estimators. Ideally, one would like to remove the bias from the variance reduction. However, this is difficult because often neither sign nor magnitude of the bias can be inferred from the variance reduction method. In the case of the cIPS estimator, however, Proposition \ref{prop:bias_cIPS} showed that the bias introduced is always negative (assuming non-negative rewards). This begs the question whether we can exploit this property to find a less bias-inducing variance-reduction method for off-policy estimation. In the following section we introduce a somewhat na\"{i}ve, yet effective, method to do so.

\section{Two-sided double clipping}

We define the \emph{two-sided double-clipping IPS} (dcIPS) estimator as
\begin{equation} \label{eq:dcIPS}
\hat{R}_{\text{dcIPS}}(\pi, {\color{ACMRed}U}, {\color{ACMDarkBlue}L}) = \frac{1}{n} \sum_{i=1}^n r(x_i, y_i) {\color{ACMDarkBlue}\max \Big\{}{\color{ACMRed} \min\{ }w(x_i, y_i), {{\color{ACMRed}U \}}} {\color{ACMDarkBlue}, \frac{1}{L}\Big\}},
\end{equation}
where $\color{ACMRed}U \geq 1$ is the upper clipping constant and $\color{ACMDarkBlue}L \geq 1$ is the lower clipping constant. The dcIPS subsumes the cIPS estimator; both estimators are equivalent for $L \rightarrow \infty$. On the other extreme, for both clipping constants approaching $1$, the dcIPS estimator converges to the mean of rewards logged in the data set:
\begin{equation} \label{eq:dcIPSLimit}
\mathbb{E}\left[\hat{R}_{\text{dcIPS}}(\pi, {\color{ACMRed}U}, {\color{ACMDarkBlue}L})\right] \rightarrow R_\text{logging} \text{ for } {\color{ACMRed}U},  {\color{ACMDarkBlue}L} \rightarrow 1.
\end{equation}
This is illustrated in Figure \ref{fig:clipping_path}, where the dcIPS (orange line) converges to the true logging reward (gray dotted line). This allows the intuitive  interpretation of dcIPS as an estimator that regularizes towards the mean of the logging policy reward and the prior variance is determined by both clipping constants ${\color{ACMRed}U}$ and ${\color{ACMDarkBlue}L}$. Under this regularization perspective, it makes sense to shrink the weights towards a positive constant ($1$ in this case) rather than to $0$, because all weights are known to be positive~\citep{lichtenberg2019regularization}.
\begin{proposition} \label{prop:bias_dcIPS} Let $w(x, y) > 0 ~ \forall x, y$, then the bias of the dcIPS estimator with clipping constants $U$ and $L$ is given by
\begin{equation}
\label{eq:dcIPSBias}
Bias(\hat{R}_{\text{dcIPS}}(\pi, {\color{ACMRed}U}, {\color{ACMDarkBlue}L})) = 
\mathbb{E}_x\mathbb{E}_{y\sim\pi}\Bigg[ 
    \Bigg(
        \underbrace{
            \mathbf{1}_{\{w(x, y) > {\color{ACMRed}U}\}}
            \left(\frac{{\color{ACMRed}U}}{w(x, y)}-1\right)
        }_{\text{Always $\leq 0$, only depends on ${\color{ACMRed}U}$}}
        +
        \underbrace{
            \mathbf{1}_{\{w(x, y) {\color{ACMDarkBlue}L} < 1\}}
            \left(\frac{1}{w(x, y){\color{ACMDarkBlue}L}}-1\right)
        }_{\text{Always $\geq$ 0, only depends on ${\color{ACMDarkBlue}L}$}}
    \Bigg)
    \underbrace{
        \mathbb{E}_r[r(x, y) | x, y]
    }_{\text{Expected reward}}
\Bigg] .
\end{equation}
\end{proposition}
\noindent Equation \ref{eq:dcIPSBias} shows that the two clipping constants contribute separately, and in opposing directions, to the overall bias of the dcIPS estimator. In other words, we can try to tune the lower clipping constant ${\color{ACMDarkBlue}L}$ so as to compensate the bias introduced by the upper clipping constant ${\color{ACMRed}U}$.


\section{Off-policy evaluation experiments} \label{sec:exp}
The synthetic experiments demonstrate that dcIPS is able to compensate the bias introduced by cIPS and can lead to lower estimation errors overall.
We used a synthetic data setting (explained in detail in the Appendix), where we collect logging data $\mathcal{D}$ from a linear stochastic logging policy that plays a multi-armed bandit environment for $n=300$ rounds. Based on $\mathcal{D}$, we estimate the expected reward of a new target policy using clipped IPS evaluators  with different clipping constants. For dcIPS, we choose the  heuristic to move ${\color{ACMRed}U}$ and  ${\color{ACMDarkBlue}L}$ in unison (i.e.,  becoming a single hyper-parameter), but more sophisticated methods to select ${\color{ACMRed}U}$ and ${\color{ACMDarkBlue}L}$ should be investigated.
We show the distribution of reward estimates (Fig. \ref{fig:clipping_path}) and estimation error components (Fig. \ref{fig:mse}) as a function of the clipping constants. The figures are best interpreted in conjunction and going from right to left on the x-axis: for large ($U = L = 100$), both cIPS and dcIPS are basically equivalent to the unclipped IPS estimator: they are unbiased but show high variance. As the clipping constants decrease, the variances of both estimates (dashed lines in \ref{fig:mse}) decreases monotonically, whereas the biases (dotted lines in \ref{fig:mse}) increase. The lower clipping of the dcIPS compensates some of the large bias suffered by the cIPS evaluator (for a given point on the x-axis, both estimators use the same upper clipping constant $U$ and thus the difference in biases reflects the bias compensation from using lower clipping as well). Thanks to this bias compensation, the dcIPS evaluator leads to lower MSE overall (solid lines in \ref{fig:mse}).


\section{Discussion and outlook} \label{sec:dis}
We analyze the bias of the clipped IPS estimator and find that negative bias provides potential for less-biased variance reduction techniques. We propose a simple method, doubly-clipped IPS, that can compensate the bias of single clipping.

One limitation is that we lack a mechanism to select clipping constants. We plan to study algorithms to select clipping constants for dcIPS in a data-driven way~\citep{bembom2008data, su2020adaptive, udagawa2023policy} and investigate theoretically when the bias of double clipping is less than standard clipping.


\begin{acks}
We thank 3 anonymous reviewers for their correction of a false statement and their useful suggestions. We also thank Yannik Stein, Vito Bellini, Matej Jakimov, Thorsten Joachims, and Harrie Oosterhuis for fruitful discussion and valuable feedback given in the context of an early talk about this project.
\end{acks}

\bibliographystyle{ACM-Reference-Format}
\bibliography{main}

\appendix

\section{Experimental setup}

We used the following classic synthetic data setting for the off-policy evaluation experiments in Section \ref{sec:exp}. 

We started by collected a logging data set $\mathcal{D}$ from a linear stochastic logging policy that played a multi-armed bandit environment for $n=300$ rounds. More specifically, the environment had $|\mathcal{Y}| = 8$ actions. Each action $y_j$ was represented by a contextual feature vector $\phi_j \in \mathbb{R}^8$, which was drawn from a normal distribution with mean equal to the 1-hot encoding of the action $y_j$ (i.e., $\phi_j[k] = 0$ in all positions $k = [1, ..., 8], k \neq j$ and $\phi_j[k] = 1$ for $k = j$), and standard deviation $\sigma = 1.0$. (Thus, context is defined implicitly via the action features.) 

The logging policy was a linear policy with weights $\beta_{\text{logging}} = [\frac{1}{9}, \frac{2}{9}, \frac{3}{9}, \dots, \frac{8}{9}]^T$ and selected actions by sampling from a softmax distribution over the scores $\Phi \beta_{\text{logging}}$, where $\Phi$ is "feature matrix" obtained by concatenating the feature vectors $\phi_j$ for all actions. 

The target policy was a similar linear policy but with ``flipped'' weights $\beta_{\text{target}} = [\frac{8}{9}, \frac{7}{9}, \frac{6}{9}, \dots, \frac{1}{9}]^T$. 

This lead to a situation where both logging and target policy had full support (all actions had a positive probability of being selected) but the logging policy ``favored'' actions in order $a_8$, $a_7$, $a_6$, ..., $a_1$, whereas the target policy  ``favored'' actions in order $a_1$, $a_2$, $a_3$, ..., $a_8$.

The reward function was again based on a linear function of the action features with weight vector  $\beta_{\text{reward}} = [0, 0.5, 0. 0.5, 0, 0, 0, 0]$ and providing a reward of $1$ for action $j$ if $\phi_j^T\beta_{\text{reward}} > 0$ and $0$ reward otherwise. (This reward is stochastic because the action features are stochastic.) Note that this reward function lead to a higher expected reward for the target policy compared to the logging policy (as can be seen in Figure \ref{fig:clipping_path}).

The data set $\mathcal{D} = \{(x_i, y_i, r_i, \pi_0(\cdot|x_i)\}, i = 1, \dots 300$ collected by the logging policy was then used to estimate the target policy reward using different IPS estimators.

\section{Proof of bias of cIPS (Proposition \ref{prop:bias_cIPS})}
\begin{proof}
For the proof we widely follow the bias expressions in \cite{dudik2011doubly, su2019cab}.
We recall the full support assumption $w(x, y) > 0 ~ \forall x, y$ (also termed \emph{absolute continuity}). 
In all steps of the proof we use the full support assumption to avoid division by $0$. 
Recall the expression for cIPS:
$\hat{R}_{\text{cIPS}}(\pi, U)= \frac{1}{n} \sum_{i=1}^n r(x_i, y_i) { \min\{ }w(x_i, y_i), {{U \}}}$.
We restrict our proof to one single sample (i.e., $n=1$). The application to the average follows from the linearity of the expectation. 
Our proof is based on the idea that we can decompose the event 
$$
{ \min\{ }w(x, y), {{U \}}} = \mathbf{1}_{\{w(x, y) > U\}} U + \mathbf{1}_{\{w(x, y) < U\}} w(x, y).
$$
In the same spirit we can rewrite
$$
w(x, y) = \mathbf{1}_{\{w(x, y) > U\}} w(x, y) + \mathbf{1}_{\{w(x, y) < U\}} w(x, y).
$$
Based on the two formulations above we get 
\begin{eqnarray*}
    \text{Bias}(\hat{R}_{\text{cIPS}}(\pi, U))
&=& \mathbb{E}[r(x, y) { \min\{ }w(x, y), {{U \}}} - r(x, y) w(x, y) ], \\
&=& \mathbb{E}[r(x, y) \mathbf{1}_{\{w(x, y) > U\}} (U - w(x,y))], \\
&=& \mathbb{E}\left[r(x, y) \mathbf{1}_{\{w(x, y) > U\}} w(x,y) \left(\frac{U}{w(x,y)} - 1 \right)\right].
\end{eqnarray*}
The bias is determined by the weights for which the upper clipping constant $U$ is exceeded. 
Now, by rearranging the terms and applying the expectation to the reward we get
\begin{eqnarray*} 
    \text{Bias}(\hat{R}_{\text{cIPS}}(\pi, U)) &=& \mathbb{E}_x\mathbb{E}_{y\sim\pi_0}\Bigl[\vphantom{\frac{U}{w(x, y)}}\mathbf{1}_{\{w(x, y) > U\}}w(x, y) 
    \left(\frac{U}{w(x, y)}-1\right)
    \vphantom{\frac{U}{w(x, y)}}\mathbb{E}_r[r(x, y) | x, y]\Bigr], \\
    &=& \mathbb{E}_x\mathbb{E}_{y\sim\pi_0}\Bigl[\vphantom{\frac{U}{w(x, y)}}\mathbf{1}_{\{w(x, y) > U\}} \frac{\pi(y|x)}{\pi_0(y|x)} 
    \left(\frac{U}{w(x, y)}-1\right)
    \vphantom{\frac{U}{w(x, y)}}\mathbb{E}_r[r(x, y) | x, y]\Bigr], \\
    &=& \mathbb{E}_x\mathbb{E}_{y\sim\pi}\Bigl[\vphantom{\frac{U}{w(x, y)}}\mathbf{1}_{\{w(x, y) > U\}} 
    \left(\frac{U}{w(x, y)}-1\right)
    \vphantom{\frac{U}{w(x, y)}}\mathbb{E}_r[r(x, y) | x, y]\Bigr].
\end{eqnarray*}
In the last line we used the importance sampling identity, i.e.,
$\mathbb{E}_{y\sim\pi_0}[f(y) \frac{\pi(y|x)}{\pi_0(y|x)}] = \mathbb{E}_{y\sim\pi}[f(y)] $, for some function $f$ of $y$,
which concludes our proof. 
\end{proof}

\section{Proof of bias of dcIPS (\ref{prop:bias_dcIPS})} 
\begin{proof}
First note that we assume $U > \frac{1}{L}$; i.e., the upper clipping constant always needs to be larger than the lower clipping constant. Note that in all steps of the proof we use the full support assumption to avoid division by $0$.  We will again focus on the expectation of a single sample, the application to the average is immediate. In the same spirit of the previous proof we rewrite 
$$
\max \Big\{{ \min\{ }w(x_i, y_i), {{U \}}} {, \frac{1}{L}\Big\}} = 
\mathbf{1}_{\{w(x, y) > U\}} U + \mathbf{1}_{ \{ 1/L < w(x, y) < U\}} w(x, y) + \mathbf{1}_{\{ 1/L > w(x, y)\}} 1/L,
$$
i.e., we clip the weights $w(x, y)$ if they fall outside the interval $[1/L, U]$. 
Again, we apply the same reasoning to the unclipped weights and get
$$
w(x, y) = 
\mathbf{1}_{\{w(x, y) > U\}} w(x, y) + \mathbf{1}_{ \{ 1/L < w(x, y) < U\}} w(x, y) + \mathbf{1}_{\{ 1/L > w(x, y)\}} w(x, y).
$$
Now, we apply this to the computation of the bias of dcIPS and get
\begin{eqnarray*}
Bias(\hat{R}_{\text{dcIPS}}(\pi, {\color{ACMRed}U}, {\color{ACMDarkBlue}L}))
&= & \mathbb{E}\left[ r(x,y) \left( \max \{{ \min\{ }w(x_i, y_i), {{U \}}} {, \frac{1}{L}\}} - w(x,y) \right) \right] \\
&=& \mathbb{E}\left[ r(x,y) \left( \mathbf{1}_{\{w(x, y) > U\}} (U -w(x,y)) + \mathbf{1}_{\{ 1/L > w(x, y)\}} (1/L - w(x,y)) \right) \right]. 
\end{eqnarray*}
Rearranging terms, applying the expectation to the reward and factoring out $w(x,y)$ then yields
\begin{eqnarray*}
Bias(\hat{R}_{\text{dcIPS}}(\pi, {\color{ACMRed}U}, {\color{ACMDarkBlue}L})) &=& 
\mathbb{E}_x\mathbb{E}_{y\sim\pi_0}\Bigg[w(x, y)
    \Bigg(
            \mathbf{1}_{\{w(x, y) > {\color{ACMRed}U}\}}
            \left(\frac{{\color{ACMRed}U}}{w(x, y)}-1\right) +
            \mathbf{1}_{\{w(x, y) {\color{ACMDarkBlue}L} < 1\}}
            \left(\frac{1}{w(x, y){\color{ACMDarkBlue}L}}-1\right)
    \Bigg)
        \mathbb{E}_r[r(x, y) | x, y]
\Bigg], \\
&=& 
\mathbb{E}_x\mathbb{E}_{y\sim\pi}\Bigg[
    \Bigg(
            \mathbf{1}_{\{w(x, y) > {\color{ACMRed}U}\}}
            \left(\frac{{\color{ACMRed}U}}{w(x, y)}-1\right) +
            \mathbf{1}_{\{w(x, y) {\color{ACMDarkBlue}L} < 1\}}
            \left(\frac{1}{w(x, y){\color{ACMDarkBlue}L}}-1\right)
    \Bigg)
        \mathbb{E}_r[r(x, y) | x, y]
\Bigg],
\end{eqnarray*}
where, as before, we used the importance sampling identity to rewrite the expectation with respect to the target distribution,  
which completes the proof. 
\end{proof}


\end{document}